\documentclass{article}
\usepackage{spconf,amsmath,graphicx}

\title{SpotFast Networks with memory augmented \\ lateral transformers   for lipreading}
\name{Peratham Wiriyathammabhum}
\address{Department of Computer Science, University of Maryland, College Park, MD, 20742, USA}
\begin{document}
%
\maketitle
\begin{abstract}
This paper presents a novel deep learning architecture for word-level lipreading. Previous works suggest a potential for incorporating a pretrained deep 3D Convolutional Neural Networks as a front-end feature extractor. We introduce a SpotFast networks, a variant of the state-of-the-art SlowFast networks for action recognition, which utilizes a temporal window as a spot pathway and all frames as a fast pathway. We further incorporate memory augmented lateral transformers to learn sequential features for classification. We evaluate the proposed model on the LRW dataset. The experiments show that our proposed model outperforms various state-of-the-art models and incorporating the memory augmented lateral transformers makes a $3.7\%$ improvement to the SpotFast networks.
\end{abstract}
\section{Introduction}
Lipreading or visual speech recognition is an interesting ability to recognize words from lip movements representing phonemes. Those lip movements are also named as visual speeches/sounds or visemes \cite{taylor2012dynamic, bear2017phoneme}. Visemes for different letters such as `p' and `b' can be very similar due to McGurk effect \cite{mcgurk1976hearing}. These letters are called homophones which are ambiguous from visual cues but can be disambiguated using additional language cues such as neighboring characters. Lipreading has many beneficial real world applications such as surveillance or assistive systems. This paper focuses on word-level automatic lipreading where the system tries to recognize a word being said given only a video sequence of moving lips without audio. 

The standard lipreading system pipeline includes mouth region cropping, mouth region compression and sequence modelling. Recently, there has been significant improvements in automatic lipreading based on deep learning systems. The current state-of-the-art \cite{weng2019learning, petridis2018end} consists of a 3D Convolutional Neural Networks (3DCNNs) as a front-end feature extractor and a 2-layered bidirectional Long-short Term Memory (biLSTM) as a back-end classifier. The system classifies the moving lips into 1-of-n words in a multi-class classification setting. Another approach \cite{Zhang_2019_ICCV} uses a transformer as a back-end instead of a biLSTM. 

\begin{figure}[t] \label{fig1}
\begin{center}
   \includegraphics[width=1.00\linewidth]{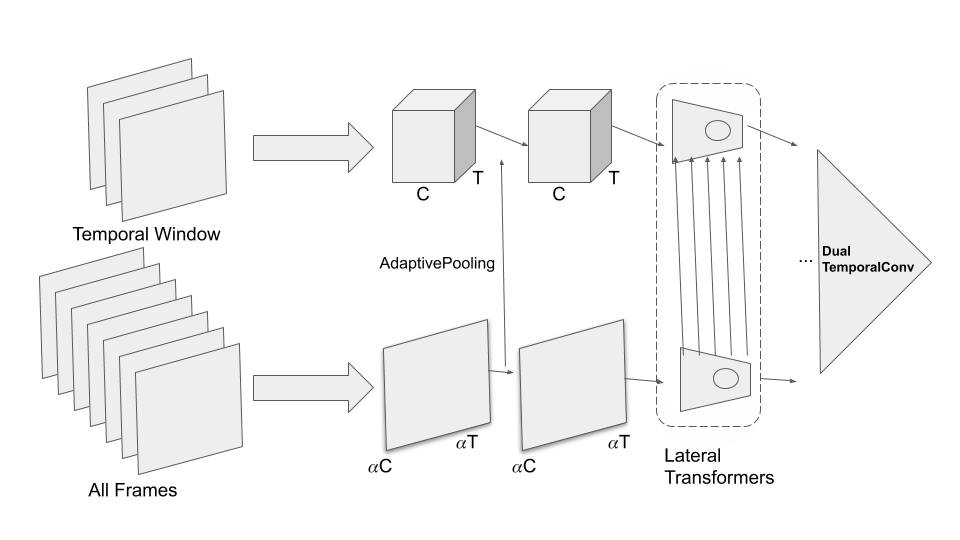}
\end{center}
\vspace{-1\baselineskip}
   \caption{\textbf{SpotFast networks with lateral transformers} utilize two pathways, consisting of a temporal window pathway and an all frame pathway. The temporal window pathway models fast evolving actions. The all frame pathway models the whole video which can incorporate more temporal contexts into the system. The adaptive average pooling layer reshapes the pre-fused features. The dual temporal convolution fuse both streams temporally and output final word-classes prediction probabilities. The lateral transformers  further model sequential information.}
\label{fig:long}
\label{fig:onecol}
\vspace{-1\baselineskip}
\end{figure}

We combine the strength from both approaches. First, we use a varient of the state-of-the-art 3DCNNs, SpotFast networks, which we replace the slow pathway of the SlowFast networks \cite{feichtenhofer2019slowfast} with the spot pathway (temporal window) to model fast lip gestures along with a fixed spotted estimation of their temporal viseme boundaries. We also utilize a pretrained SlowFast model convolutional parameters from the Kinetics-400 dataset \cite{carreira2017quo} by introducing an additional adaptive average pooling layer to reshape the convolutional feature from the all frame pathway into a fixed temporal window size of the spot pathway. We additionally put a dual temporal convolution layer on top of the networks to temporally convolve each pathway with different kernel sizes then fuse them into word-classes prediction. By using this dual temporal convolution back-end, we observe a more stabilize training empirically. 

Second, we introduce memory augmented lateral transformers to further model temporal sequential information from the feature set extracted from convolutional layers of the SpotFast networks. Lateral transformers are two vanilla transformers with a lateral connection after each hidden layer which fuse features from the all frame pathway to the temporal window pathway. We also augment the product-key memory \cite{lample2019large} to the layer before the last of each pathway to increase the model capacity and stability. Utilizing the three stage training procedure \cite{petridis2017end}, we achieve the new state-of-the-art for word-level lipreading using only RGB/grayscale input.

The contributions of this paper are (i) We propose a novel state-of-the-art deep learning architecture for end-to-end word-level lipreading. (ii) We evaluate the effect of the temporal window sizes on our proposed SpotFast networks. 

\section{Related Works}
\textbf{Pretrained video models.} The state-of-the-art action recognition models are primary choices when we want to perform transfer learning and finetune video models for other tasks. Lipreading in the deep learning era incorporates recent advancements in action recognition such as C3D (3D convolution) \cite{tran2015learning} or I3D (deep 3DCNNs) \cite{carreira2017quo} as a better front-end and achieves state-of-the-art \cite{stafylakis2017combining, weng2019learning}. We further incorporate another recent advancement in action recognition, SlowFast networks \cite{feichtenhofer2019slowfast}, as our front-end feature extractor. 

\hspace{-1.2\baselineskip}\textbf{Transformers.} As a recent promising alternative to biLSTMs and biGRUs in sequence modelling, a transformer model \cite{vaswani2017attention} is widely deployed in many state-of-the-art systems in various NLP tasks. A transformer consists of a stack of multi-head self attention and feed forward modules in an autoencoder setting. There is no recurrence in the model so training a transformer can be easily paralleled. Lipreading also incorporates the transformer model and makes substantial improvements on many datasets \cite{afouras2018deep, Zhang_2019_ICCV}. We follow this direction and propose a lateral transformer for sequence modelling in our SlowFast-based architecture.

\hspace{-1.2\baselineskip}\textbf{Memory.} Incorporating memory is an approach to increase the capacity of a neural networks without increasing too much computation. Those memory augmented neural networks can be an efficient and effective way to represent variable length inputs for question answering \cite{weston2014memory, sukhbaatar2015end} or learning with limited data for few-shot learning \cite{santoro2016meta, zhu2018compound}. 
A recently proposed product-key memory \cite{lample2019large} is a promising neural network layer which can be incorporated into transformer-based models and greatly increase the capacity with only half computation. The memory holds a table of key-value entries with a multi-head mechanism which can be trained end-to-end. Each memory head has its own query networks and sub-keys but shares the same values with other heads. For each head, an input query will be compared to all keys in a nearest neighbor setting and the sparse weighted-sum of the corresponding memory values of the $k$ nearest keys will be the output. The output of the memory layer will be the sum of all outputs from all heads.

\section{Proposed Method}

\begin{figure}[t] 
\vspace{-0.5\baselineskip}
\begin{center}
   \includegraphics[width=1.00\linewidth]{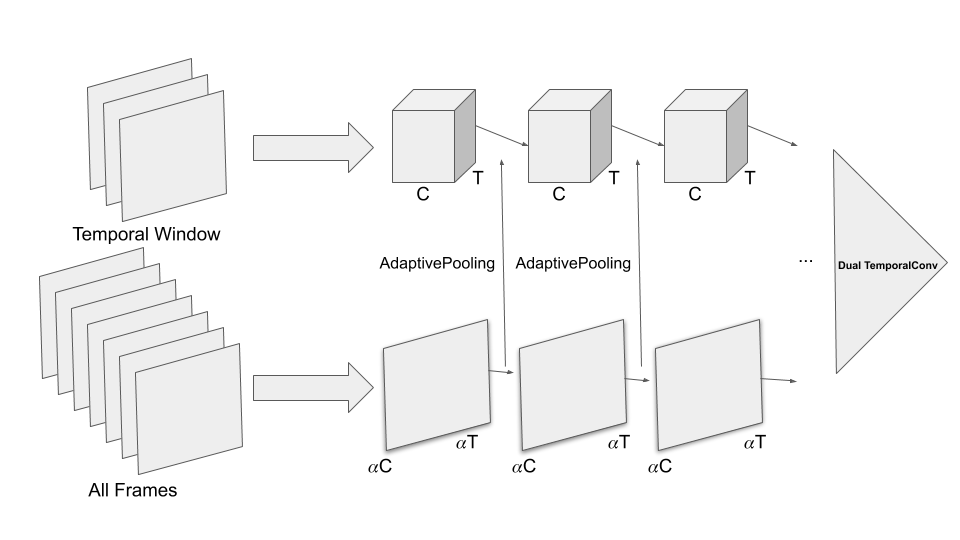}
\end{center}
\vspace{-1.5\baselineskip}
   \caption{\textbf{SpotFast networks}}
   \label{sys_overview}
\label{fig:long}
\label{fig:onecol}
\vspace{-1\baselineskip}
\end{figure}

\subsection{SpotFast Networks}
SlowFast networks \cite{feichtenhofer2019slowfast} consist of two pathways, slow and fast. The slow pathway is not suitable for lipreading since lip movements are fast evolving actions. We use the knowledge that the target word to be lipreaded for word-level lipreading is always keyword spotted. We then propose an alternative spot pathway to the slow pathway such that the fast evolving action will be captured by the networks. The spot pathway is a temporal window centered at the keyword-spotted frame. We keep the fast pathway as all frames which is the same as the original SlowFast networks. To fuse fast pathway to spot pathway via lateral connections, we use convolution fusion as in the original SlowFast networks with additional adaptive average pooling to temporally reshape the features from all frames into a fixed length temporal window. We use the SpotFast networks as a front-end to extract spatio-temporal features.

\begin{figure}[t] 
\vspace{0.5\baselineskip}
\begin{center}
   \includegraphics[width=0.5\linewidth]{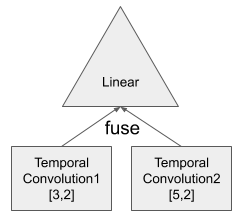}
\end{center}
\vspace{-1.5\baselineskip}
   \caption{\textbf{Dual Temporal Convolutions}}
   \label{sys_overview}
\label{fig:long}
\label{fig:onecol}
\vspace{-1\baselineskip}
\end{figure}

\subsection{Dual 1D Temporal Convolution Networks (TC)}
For the back-end, we deploy a simple two-layered 1D temporal convolution to aggregate temporal information on each pathway. The output from each pathway of the SpotFast networks is average-pooled to 3 modes, \textit{[batch\_size, feature\_size, time\_step]}. The first layer doubles the in\_channel using a filter with a kernel size of 3 and a stride of 2 (temporal window) or a kernel size of 5 and a stride of 2 (all frames). Then, we apply a sequence of batchnorm, ReLU activation and max pooling with a kernel size of 2 and a stride of 2. The second layer has the same parameters as the first layer while doubles the out\_channel from the first layer to quadruple the initial in\_channel. Next, the features from both pathways are averaged-pooled and fused using concatenation. Lastly, a linear feed forward layer maps the fused feature into word-class probabilities.

\begin{figure}[t] 
\begin{center}
   \includegraphics[width=1.00\linewidth]{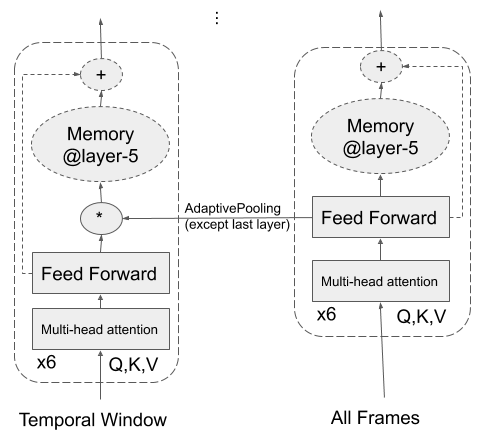}
\end{center}
\vspace{-1.5\baselineskip}
   \caption{\textbf{Memory Augmented Lateral Transformers}}
   \label{sys_overview}
\label{fig:long}
\label{fig:onecol}
\vspace{-1\baselineskip}
\end{figure}

\subsection{Memory Augmented Lateral Transformers}
To increase the capacity of the back-end, we put a transformer encoder on top of each pathway of the SpotFast networks to further learn features for classification. For each pathway, we use a 6-layered transformer encoder (base model) consisting of a multi-head attention and a feed forward module which maps the input features in an autoencoder setting. We augment each transformer with the product-key memory at the layer before the last as in \cite{lample2019large} (layer 5 for the base transformer model) to increase the capacity and stabilize training. The output feature after the feed forward layer is fed into the memory and is added to the output of the memory module via a skip connection. However, we empirically observe that independently putting the transformer on top of each pathway makes learning difficult. The loss does not going down in a reasonable rate or even diverged. To solve this problem, we add lateral connections to all layers (except the last layer) of both transformer encoders from the all frame pathway to the temporal window pathway. 

\subsection{Implementation details}
We use the Kinetics-400 pretrained model from the 8x8 ResNet-50 SlowFast networks with non-local blocks. The temporal window pathway has the feature size of $2048$. The all frame pathway has the feature size of $256$. The outputs of both pathway are average-pooled with a kernel size of $[1,4,4]$ and a stride of $1$. 

The transformer encoder has $8$ heads. The positional encoders for the same position in both pathway encode with the same values by using offset index added to the temporal window. The positional encoders has a dropout of $0.1$. The lateral connections are consisted of $1x1x1$ convolutions with padding of $1$ and an adaptive average pooling in the temporal dimension for the all frame pathway. Then, the features from the temporal window pathway and the lateral connection are fused by concatenation followed by a sequence of linear feed forward, batchnorm and ReLU activation. 

The product-key memory has $4$ heads. The vectors are of $128$ dimensions. The memory uses $32$ nearest neighbors. We use query batchnorm and a value dropout of $0.1$. The temporal window pathway has $168^2$ values and the all frame pathway has $50^2$ values. We apply a layernorm to the output of the memory. 

Our implementation is based on the PyTorch library \cite{paszke2019pytorch}. All models are trained using 4 NVIDIA P6000 GPUs and 16 CPUs. The networks are trained using the Adam optimiser \cite{kingma2014adam} and label smoothed cross entropy loss with the smoothing parameter of $0.1$. In the first phrase, we train the SpotFast networks end-to-end using an initial learning rate of $2.5e-4$, a weight decay of $1e-4$ and a batch size of $84$ for $10$ epochs. In the second phase, we fixed the parameters of the SpotFast networks and train only the lateral transformers and the dual temporal convolutions using an initial learning rate of $2.25e-4$, a weight decay of $3e-4$ and a batch size of $84$ for $5$ epochs. In the third phrase, we finetune the whole network end-to-end using an initial learning rate of $1.566e-4$, a weight decay of $1e-4$ and a batch size of $64$. We use the cosine annealing scheduler with restart \cite{loshchilov2016sgdr} where the parameters are $T_0=5$, $T_{mul}=1$ and $eta\_min=0$ for all phrases. We use a linear warmup with $2000$ steps for the first phrase and $1000$ steps for the second and the third phrases. For the third phrase, we chain the ReduceLR scheduler in addition to the cosine scheduler and reduce the learning rate by a factor of $2$ when the validation loss does not decrease. The third phrase takes $30$ epochs in total.

The codebase along with the pretrained models will be released to foster research in the community. 

\section{Experiments}
\subsection{Dataset, preprocessing and augmentations}
We conduct the experiments using the Lip Reading in the Wild (LRW) dataset \cite{chung2016lip}. The LRW dataset consists of 488,766 training, 25,000 validation and 25,000 testing short video clips. The video clips are talking face videos extracted from BBC TV broadcasts. There are 500 target words in this dataset. For each video clip, there are 29 frames. We use RGB frames in our experiments because the pretrained Kinetics models are in RGB. The dataset contains words with similar visemes such as `SPEND' and `SPENT' (tenses) or `BENEFIT' and `BENEFITS' (plural forms). The words are not totally isolated and the word boundaries are not given. Those words are surrounded by irrelevant parts of the utterances which may provide contexts or are just noises.  

The preprocessing consists of cropping mouth regions using fixed coordinates since the LRW dataset is already centered spatially. We then normalize all frames into $[0,1]$, subtract them with $0.45$ and divide them with $0.225$ which are preprocessing parameters from the Kinetics pretrained models.

We perform data augmentations like other previous works. The augmentations are done during training consisting of random upsampling ($[122, 146]$), random crop (crop to $112\times112$ pixels) and random horizontal flip (with a probability of $0.5$). We augment the same way for every frame in both pathways. During validation and testing phase, we upsampling to $122\times122$ pixels and uniform crop to $112\times112$ pixels.

\subsection{Comparisons with the state-of-the-arts}
We summarize the state-of-the-art methods in Table.\ref{stable1}. Our final SpotFast with memory-augmented lateral transformers outperform previous state-of-the-art methods, including the one using optical flow information. Our proposed method makes an improvement over the prior state-of-the-art \cite{weng2019learning} by $0.3\%$ and the prior RGB/grayscale state-of-the-art \cite{Zhang_2019_ICCV} by $0.7\%$.

\begin{table}[] \label{stable1}
\centering
\caption{Top-1 test accuracies of the state-of-the-art methods on the LRW dataset.}
\vspace{0.5\baselineskip}
\label{stable1}
\begin{tabular}{l|c}
\hline
 Method    & Accuracy  \\
\hline
 LRW \cite{chung2016lip} & 61.1  \\
WAS \cite{chung2017lip} & 76.2   \\
ResNet+biLSTMs \cite{stafylakis2017combining} & 83.0 \\
ResNet+biGRUs \cite{petridis2018end} & 83.4 \\
ResNet+focal block+transformer \cite{Zhang_2019_ICCV} & 83.7   \\ 
I3D+biLSTMs \cite{weng2019learning} & 84.1   \\ 
\hline
\textbf{SpotFast with lateral transformers} &  \textbf{84.4}  \\
\textbf{(ours)} & \\
\hline
\end{tabular}
\end{table}

\subsection{The effects of the window sizes in SpotFast Networks}
We determine the optimal temporal window size for the temporal window pathway using a grid search over a set of $3$ values, $\{15, 19, 23\}$. The validation and test accuracies are summarized in Table.\ref{stable2}. The validation and test accuracies are comparable (SpotFast front-end increases the accuracy from $74.6\%$ of ResNet+temporalConv \cite{stafylakis2017combining}.) with the window size of $23$ being the best ($6.1\%$ increase in accuracy.). We then proceed the training of the SpotFast networks to the next phase with the temporal window size of $23$. The $3$ values are heuristics from the word-boundary statistics estimated from the training partition of the LRW dataset. The word boundary distribution has the mean of $10.59$ (We round it to 11.) and the standard deviation of $3.2$ (We round it to 4.). For an approximately normal distribution, we use the 68-95-99.7 rule (three-sigma rule of thumb which follows the cumulative distribution function of the normal distribution. It will become 0-75-89 by Chebyshev's inequality for any distributions.) to create $3$ values where the bands will cover 68-95-99.7 percents of the data population using $\{\mu+\sigma, \mu+2\sigma, \mu+3\sigma\}$ which becomes $\{15, 19, 23\}$. We can also observe from Table.\ref{stable2} that incorporating the lateral transformers into SpotFast networks and training the whole system end-to-end can further increase the test accuracy upto $3.7\%$.

\begin{table}[] \label{stable2}
\vspace{0.5\baselineskip}
\centering
\caption{The top-1 accuracies of the SpotFast networks (without lateral transformers) varying the temporal window sizes.}
\vspace{0.5\baselineskip}
\label{stable2}
\begin{tabular}{l|c|c}
\hline
 Temporal window size    & Validation & Test  \\
\hline
 SpotFast-15 & 81.3 & 80.6\\
\hline
SpotFast-19 & 81.2  & 80.4 \\ 
\hline
\textbf{SpotFast-23} &  \textbf{81.5} & \textbf{80.7}\\
\hline
\end{tabular}
\end{table}

\begin{figure}[t] \label{figstat}
\begin{center}
   \includegraphics[width=1.00\linewidth]{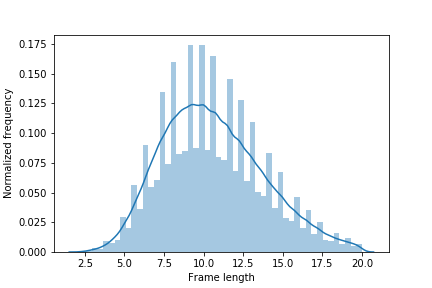}
\end{center}
\vspace{-1.5\baselineskip}
   \caption{\textbf{Word boundary distribution of the LRW training set in number of frames.}}
\label{fig:long}
\label{fig:onecol}
\vspace{-1\baselineskip}
\end{figure}




\section{Summary}
We propose a SpotFast networks with lateral transformers for word-level lipreading. We utilize a two-stream deep 3DCNNs, a temporal window pathway and an all frame pathway. We evaluate a heuristics based on the 68-95-99.7 rule to select the optimal temporal window size. We show that incorporating lateral transformer can improve the accuracy for $3.7\%$. We also show that our proposed model can outperform various state-of-the-art models. Some possible future directions include incorporating an optical flow input to the model and increasing the augmented-memory capacity.

\bibliographystyle{IEEEbib}
\bibliography{egbib}

\end{document}